\documentclass[conference]{IEEEtran}

\usepackage{cite}

\usepackage[flushleft]{threeparttable} 
\usepackage{booktabs}

\usepackage{graphicx}  
\usepackage[caption=false,font=scriptsize]{subfig}  
 
\usepackage{amsmath,amssymb,amsfonts}   
\usepackage{mathtools}
\usepackage{algorithm}
\usepackage{algorithmic}
\usepackage{textcomp}
\usepackage{xcolor}
\usepackage{array} 
\usepackage{multirow} 
\usepackage{enumerate} 

\usepackage{url}

\usepackage{url}
\usepackage{float}
\usepackage{array}  
\usepackage{enumitem} 

\makeatletter
\newcommand*\titleheader[1]{\gdef\@titleheader{#1}}
\AtBeginDocument{%
  \let\st@red@title\@title
  \def\@title{%
    \bgroup\normalfont\large\centering\@titleheader\par\egroup
    \vskip1.5em\st@red@title}
}
\makeatother

\title{Learning More with Less: A Generalizable, Self-Supervised Framework for Privacy-Preserving Capacity Estimation with EV Charging Data}
\titleheader{This manuscript has been accepted in IEEE Transactions on Industrial Informatics. \\ DOI: \textbf{https://doi.org/10.1109/TII.2025.3613385} \\ \vspace{0.5cm} \scriptsize \copyright~2025 IEEE.  Personal use of this material is permitted.  Permission from IEEE must be obtained for all other uses, in any current or future media, including reprinting/republishing this material for advertising or promotional purposes, creating new collective works, for resale or redistribution to servers or lists, or reuse of any copyrighted component of this work in other works.
}

\author{
  \IEEEauthorblockN{Anushiya Arunan}
  \IEEEauthorblockA{
    Engineering Product Development\\
    Singapore University of Technology and Design\\
    Email: anushiya\_arunan@mymail.sutd.edu.sg
  }
  \and
  \IEEEauthorblockN{Yan Qin}
  \IEEEauthorblockA{
    School of Automation\\
    Chongqing University\\
    Email: yan.qin@cqu.edu.cn
  }
  \and
  \IEEEauthorblockN{Xiaoli Li}
  \IEEEauthorblockA{
    Information Systems Technology and Design\\
    Singapore University of Technology and Design\\
    Email: xiaoli\_li@sutd.edu.sg
  }
  
    \and
  \IEEEauthorblockN{U-Xuan Tan}
  \IEEEauthorblockA{
    Engineering Product Development\\
    Singapore University of Technology and Design\\
    Email: uxuan\_tan@sutd.edu.sg
  }
  
    \and
  \IEEEauthorblockN{H. Vincent Poor}
  \IEEEauthorblockA{
    Electrical and Computer Engineering\\
    Princeton University\\
    Email: poor@princeton.edu
  }
  
    \and
  \IEEEauthorblockN{Chau Yuen}
  \IEEEauthorblockA{
    Electrical and Electronics Engineering\\
    Nanyang Technological University\\
     Email: chau.yuen@ntu.edu.sg
  }

}

\newcolumntype{M}[1]{>{\centering\arraybackslash}p{#1}}

\begin{document}


\twocolumn
\maketitle
\begin{abstract}
Accurate battery capacity estimation is key to alleviating consumer concerns about battery performance and reliability of electric vehicles (EVs). However, practical data limitations imposed by stringent privacy regulations and labeled data shortages hamper the development of generalizable capacity estimation models that remain robust to real-world data distribution shifts. While self-supervised learning can leverage unlabeled data, existing techniques are not particularly designed to learn effectively from challenging field data---let alone from privacy-friendly data, which are often less feature-rich and noisier. In this work, we propose a first-of-its-kind capacity estimation model based on self-supervised pre-training, developed on a large-scale  dataset of privacy-friendly charging data snippets from real-world EV operations. Our pre-training framework, \textit{snippet similarity-weighted masked input reconstruction}, is designed to learn rich, generalizable representations even from less feature-rich and fragmented privacy-friendly data. Our key innovation lies in harnessing contrastive learning to first capture high-level similarities among fragmented snippets that otherwise lack meaningful context. With our snippet-wise contrastive learning and subsequent similarity-weighted masked reconstruction, we are able to learn rich representations of both granular charging patterns within individual snippets and high-level associative relationships across different snippets. Bolstered by this rich representation learning, our model consistently outperforms state-of-the-art baselines, achieving 31.9\% lower test error than the best-performing benchmark, even under challenging domain-shifted settings affected by both manufacturer and age-induced distribution shifts. Source code is available at https://github.com/en-research/GenEVBattery.

\end{abstract}


\section{Introduction}
\IEEEPARstart{E}{lectric} vehicles (EVs), powered by lithium-ion batteries, possess the transformative potential to accelerate the decarbonization of the transportation sector \cite{li2022battery}. However, consumer concerns, such as uncertainties about driving range, long charging times, and battery safety, present significant barriers to the widespread adoption of EVs over their fossil fuel-based counterparts \cite{javadnejad2023analyzing}. A key to alleviating these concerns is the availability of accurate battery capacity estimation---which serves as the foundation for making informed and optimal decisions about journey planning and battery maintenance \cite{sulzer2021challenge}.

Batteries gradually lose capacity over time, with the deterioration influenced by a combination of factors such as the intensity of charge-discharge cycles \cite{zhang2024multi} and operating environments \cite{wang2022battery}. As a result, accurately estimating battery capacity is highly challenging, especially under real-world conditions outside of controlled laboratory settings. Machine learning---especially deep learning (DL)---provides a flexible approach for capturing the complex dependencies affecting capacity degradation. By leveraging data mining and pattern analysis, DL models can bypass the need for detailed mechanistic modeling to learn generalizable features \cite{11083573, wang2025itformer}. In recent years, DL techniques  have become indispensable for the battery research community, driven by the growth in benchmark battery datasets and architectural advancements in time series modeling \cite{dos2021lithium, wang2023self}. 

However, training data from actual EV operations is often scarce in practice due to two critical constraints. First, as regulatory scrutiny over responsible artificial intelligence increases, data security and privacy laws are poised to become more stringent globally \cite{sun2025sanitizable, sun2024efficient, you2025framework}. 
From the seminal General Data Protection Regulation to the newly proposed EU Artificial Intelligence Act \cite{gstrein2024general}, regulations  increasingly mandate user consent and minimization of collected data---which limit EV manufacturers' ability to curate rich training data from real-world driving operations \cite{ahmad2024comprehensive}. Second, even when battery data is available, parsing complex sensor readings and annotating them with ground-truth capacity labels is a resource-intensive endeavor requiring specialized domain expertise \cite{sulzer2021challenge}. 
With recent advancements like fast-charging batteries further complicating the analysis of battery behaviors \cite{zhou2022state} and raising data parsing costs, it is often only economically feasible to label a small portion of collected data \cite{lu2023deep}. In these  data-scarce environments, a continued reliance  on labeled data alone for model training will result in overfitted models of little practical applicability, struggling to generalize beyond the limited number of training examples \cite{wang2023self, wang2025leveraging, wang2024incorporating}.

Therefore, we posit that, for capacity estimation models to be effective under real-world data constraints, they must possess the following essential capabilities: 
\begin{enumerate}[label=\arabic*.]  
\item Achieve accurate capacity estimation even with privacy-friendly field data, which have significantly lower feature-richness and contextual information \cite{sulzer2021challenge} than laboratory-generated experimental data;  

\item Leverage the untapped potential of unlabeled data, which is more readily available and offers a greater diversity of training examples than small labeled datasets.
\end{enumerate}

\vspace{0.15cm}
However, the few battery studies from actual EV operations tend to fall short of these requirements. Typically focused on public transportation fleets rather than private vehicles, studies often rely on detailed, privacy-sensitive operational data such as on-road driving behaviors and mileage traveled \cite{she2019battery, song2020intelligent, hong2019fault, hong2021online, wang2021data} to supplement battery data and achieve accurate battery health prognostics. With growing data-privacy regulations, the granular operational data needed by existing prognostics models \cite{she2019battery, song2020intelligent, hong2019fault, hong2021online, wang2021data} is less likely to be available, especially from private EV owners.
Moreover, the dependence on supervised learning and labeled data \cite{she2019battery, song2020intelligent, deng2023prognostics, hong2019fault, lu2024towards , she2021battery, xu2021estimation, hong2021online, wang2021data} imposes a practical limit on the generalization capabilities of these models, as it is not sustainable to pre-emptively create extensive, labeled training datasets for every possible test condition.

A promising paradigm for learning generalizable models under labeled data scarcity is \textit{self-supervised} pre-training followed by supervised fine-tuning \cite{chen2020big}. Known for its success in computer vision, self-supervised pre-training involves solving auxiliary tasks---such as reconstructing masked out patches in images \cite{he2022masked} or contrastive learning of image similarities \cite{chen2020simple}---to learn generalized representations and patterns of the input data, without requiring any target labels. In particular, contrastive learning has proven capable of extracting discriminative features, even from noisy or incomplete data \cite{wu2025tcg}. By learning effectively from unlabeled data, the pre-trained model can be fine-tuned with a small amount of labeled data to adapt to any downstream task, thus reducing the massive data labeling efforts needed for fully supervised models.

Battery health prognostics stands to benefit greatly from self-supervised learning, especially given the costly domain expertise needed for data labeling. However, the few nascent efforts \cite{wang2024lithium, wang2024enhanced, li2023contrastive} to leverage self-supervised learning typically draw from computer vision techniques for images, and thus are not particularly well-suited to learn from time series battery measurements. For instance, some studies employ computationally intensive transformations to convert time series battery data to images \cite{wang2024lithium} and adopt parameter-heavy architectures like vision transformers \cite{wang2024lithium, wang2024enhanced} for the self-supervised representation learning. However, these approaches may not adequately capture important temporal dependencies in the battery data, resulting in a higher labeled data demand of at least 25-30\% for a good fine-tuning performance downstream \cite{wang2024lithium, wang2023self, wang2024enhanced}. Furthermore, self-supervised learning techniques have hardly been investigated on data from actual EV operations \cite{wang2024lithium, wang2023self, wang2024enhanced, hannan2021deep, li2023contrastive, wang2024unlocking}---let alone on more challenging privacy-friendly data. Hence, their generalization under practical data constraints and real-world data distribution shifts are unknown.

\begin{figure}[t]
    \centering
    \includegraphics[width=0.85\columnwidth]{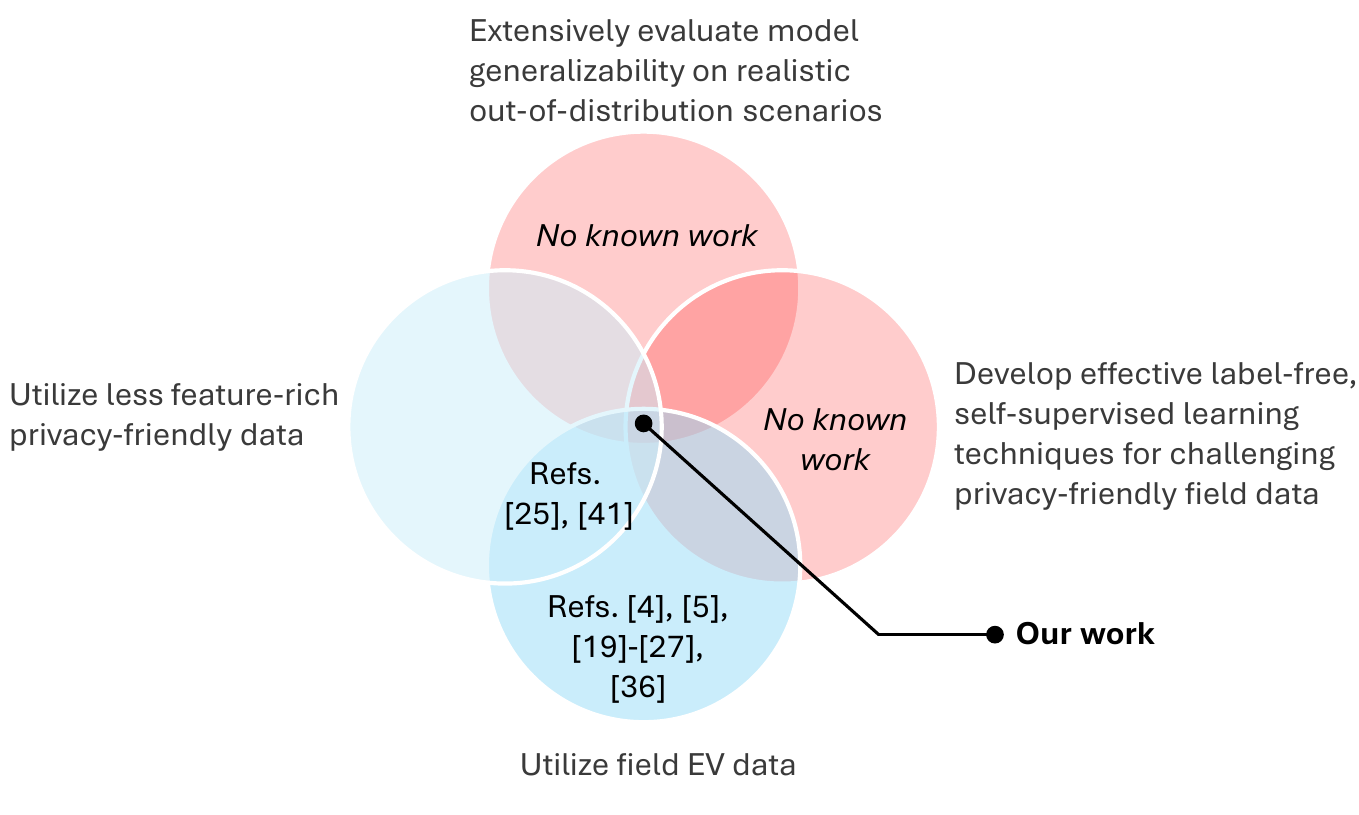} 
    \vspace{-1em}
    \caption{Contribution of our work to the domain of practical EV battery capacity estimation using field data.}
\vspace{-0.7em}
    \label{Fig_WorkNovelty}

\end{figure}

In this work, we propose a first-of-its-kind capacity estimation model based on the self-supervised pre-training paradigm, developed and tested on a large-scale dataset of nearly 200,000 privacy-friendly, short-sequenced snippets of battery monitoring data from real-world EV charging operations (Fig. \ref{Fig_WorkNovelty}). Our pre-training framework, \textit{snippet similarity-weighted masked input reconstruction}, is designed to learn rich, generalizable representations even from less feature-rich and fragmented privacy-friendly data. As a self-supervised technique, the basic principle of masked input reconstruction is to reconstruct the masked-out portions of the data as accurately as possible to the original, thereby inducing the learning of meaningful representations in the original data without any target labels. However, the higher noise-to-signal ratio in fragmented snippets of partial charging data can significantly challenge the masked reconstruction process and hamper effective representation learning.

A key innovation of our pre-training design is thus leveraging contrastive learning to learn meaningful similarity associations among the fragmented snippets, which enables the model to focus on similarities between the masked and original snippets, while suppressing noise from dissimilar snippets during the reconstruction process. Although some state-of-the-art time series pre-training frameworks \cite{lee2024learning, dong2024simmtm}  also utilize contrastive learning, their computer vision-centric approaches (e.g., hierarchical contrastive learning or reliance on multiple masked augmentations) often introduce additional noise and complexity, which impedes optimal representation learning from already challenging field datasets. In comparison, we harness contrastive learning in a simple but impactful manner to first learn high-level similarity associations among fragmented snippets that otherwise lack meaningful context. With our snippet-wise contrastive learning and subsequent similarity-weighted masked reconstruction, we can learn rich representations of  both granular charging patterns within individual snippets \textit{and} high-level associative relationships across different snippets.

Our model's generalizability is evaluated on a suite of diverse and challenging capacity estimation scenarios encountered by EV users and manufacturers---(i) adaptability to age-induced data distribution shifts across EV lifetime, (ii) transferability across different EV manufacturers, and (iii) robustness to novel test patterns unseen in the limited labeled fine-tuning data. Bolstered by the rich representations learnt by the pre-trained encoder, we consistently outperform existing state-of-the-art baselines \cite{lee2024learning,dong2024simmtm, zerveas2021transformer, eldele2024tslanet} across these out-of-distribution scenarios, despite using only 10\% labeled fine-tuning data. 

Overall, the key contributions of this work are: 
\begin{enumerate}[label=\arabic*.]  

\item As the first generalizable capacity estimation framework developed on field EV data, our work fills an important gap in existing literature:  how to learn effectively from the less feature-rich and privacy-friendly  partial charging data to still yield robust and generalizable capacity estimation.

\item Our model's generalizability extends to highly domain-shifted settings affected by both manufacturer and age-induced distribution shifts, where it still achieves 31.9\% lower test error than the best-performing benchmark.

\item Departing from existing studies, our work demonstrates that developing a generalizable model does not always require extensive labeled data. This marks a pivotal step towards practically implementable capacity estimation models, amidst the dual challenges of privacy and labeled data constraints.

\end{enumerate}

The rest of the paper is organized as follows. Section \ref{data_generation} establishes the data distributions for realistic out-of-distribution problem settings. Section \ref{methodology} presents the proposed capacity estimation framework, with a focus on the self-supervised pre-training scheme. Section \ref{results_discussion} discusses the experimental results. Section \ref{conclusion} concludes and highlights future directions.

\section{Data distribution generation for out-of-distribution problem design}
\label{data_generation}

To evaluate our capacity estimation model, we design a suite of challenging problem settings investigating diverse aspects of generalizability, including adaptability to age-induced data distribution shifts, cross-manufacturer transferability, and robustness to novel test patterns. This section establishes the data distributions for the out-of-distribution problem settings.

For the problem design, we leverage a newly released dataset of EV charging records, collected over several years from charging stations \cite{he2022evbattery}. This is the largest publicly-available field data from EVs normally operating in real-world conditions. Notably, this dataset is fundamentally privacy-friendly as it contains only charging records. Battery charging data, unlike discharging data, cannot capture sensitive information on day-to-day operational behaviours \cite{kang2020electric}. Moreover, as an additional privacy-preserving measure, the raw data is already pre-segmented, and only shorter, fragmented snippets of partial charging data are provided. While this does not introduce systematic biases or diminish the dataset’s utility \cite{he2022evbattery}, the limited privacy-friendly features being spread across fragmented snippets makes the dataset highly challenging to work with. Nonetheless, the dataset serves as a realistic standard for the minimalistic data likely to be available under stricter privacy regulations.

\begin{table}[t]
\scriptsize
\centering
\caption{Overview of generated data distributions } 
\begin{tabular}{M{1.5cm} M{1.5cm} M{3cm} M{1.1cm}}
\toprule
EV manufacturer &  Distribution name & Accumulated mileage \hspace {1cm} $m$ & Short-form  \\
\midrule
\multirow{3}{*}{EVM1} & Distribution 1 & $m \leq$ 100,000 km & EVM1\textbf{D1}\\
 & Distribution 2 & 100,000 $< m \leq$ 150,000 km & EVM1\textbf{D2}\\
\vspace{1mm}
 & Distribution 3 & $m >$ 150,000 km & EVM1\textbf{D3}\\
\bottomrule
\end{tabular}
\label{Table_Intro_3_distributions}
\vspace{-0.5em}
\end{table}

\begin{figure}[t!]
    \centering
    \begin{minipage}[t]{0.8\columnwidth}
        \centering
        \subfloat[]{\includegraphics[width=\columnwidth]{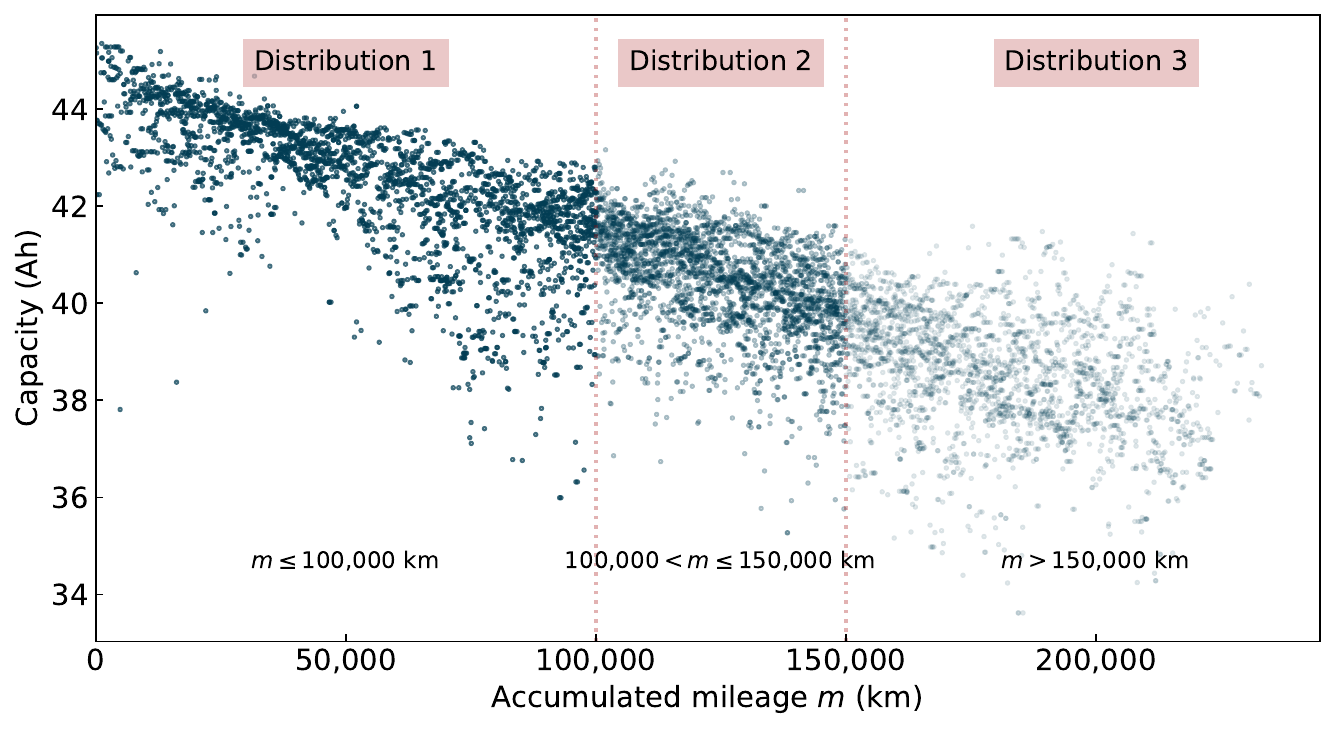} \label{scatterplot_mileage_distrib}}
    \end{minipage}
    
    \vspace{0.2cm}
    \begin{minipage}[t]{0.8\columnwidth}
        \centering
        \subfloat[]{\includegraphics[width=\columnwidth]{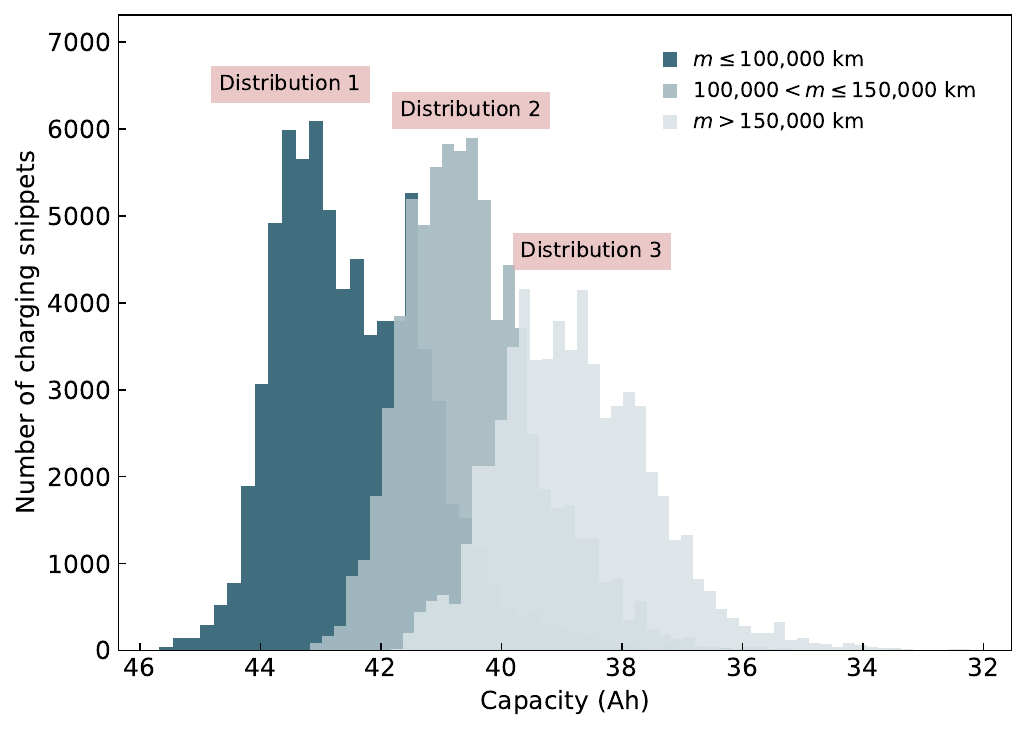} \label{histogram_capacity_distrib}}
    \end{minipage}

    \caption{Visualization of the three generated data distributions. Considering the charging snippets across all EVs, \textbf{(a)} illustrates the inverse relationship between the mileage and capacity labels of the charging snippets, particularly, highlighting the capacity degradation that occurs through distributions 1 to 3. \textbf{(b)} This capacity degradation results in the notable domain shifts.}
    \label{Fig_intro_3_distributions}

\vspace{-0.5em}
\end{figure}

The dataset contains EVs from three manufacturers, aliased as Manufacturers 1, 2, and 3. Within each manufacturer, the charging data snippets are provided at the vehicle-level. Each charging snippet only contains basic battery monitoring variables such as charging current, voltage, temperature, and state of charge, and is stored as a multivariate time series sequence. Every charging snippet has a \textit{mileage} label indicating the EV's accumulated mileage at the time of charging. However, only a subset of the snippets have \textit{capacity} labels due to the complexity of the charging processes and the resource-intensive involvement of EV manufacturer engineers for capacity labeling \cite{he2022evbattery}. 

We primarily use data from Manufacturer 1, which contains the largest number of charging snippets among the three  manufacturers. We partition the charging snippets based on their mileage labels, to generate three meaningfully different data distributions, labeled 1 to 3, demarcated by the mileage thresholds shown in Table \ref{Table_Intro_3_distributions}. Fig. \ref{Fig_intro_3_distributions}a captures the inverse relationship between the mileage and capacity labels of the charging snippets, and particularly, the capacity degradation that occurs as EVs progress from Distributions 1 to 3 over their operational lifetime. This capacity degradation results in the notable domain shifts observed among the three generated distributions (Fig. \ref{Fig_intro_3_distributions}b), which facilitates a rigorous  evaluation of our capacity estimation model's generalizability to age-induced data distribution shifts occurring over EV lifetime.

Aside from Manufacturer 1, we also strategically leverage the smaller dataset from Manufacturer 3 to evaluate additional aspects of model generalizability, such as transferability across different EV manufacturers. To maintain brevity, we discuss this problem design together with the experimental results in later sections.

\section{Proposed capacity estimation framework}\label{methodology}

\begin{figure*}[h!]
    \centering
    \subfloat[]{%
        \includegraphics[width=0.7\textwidth]{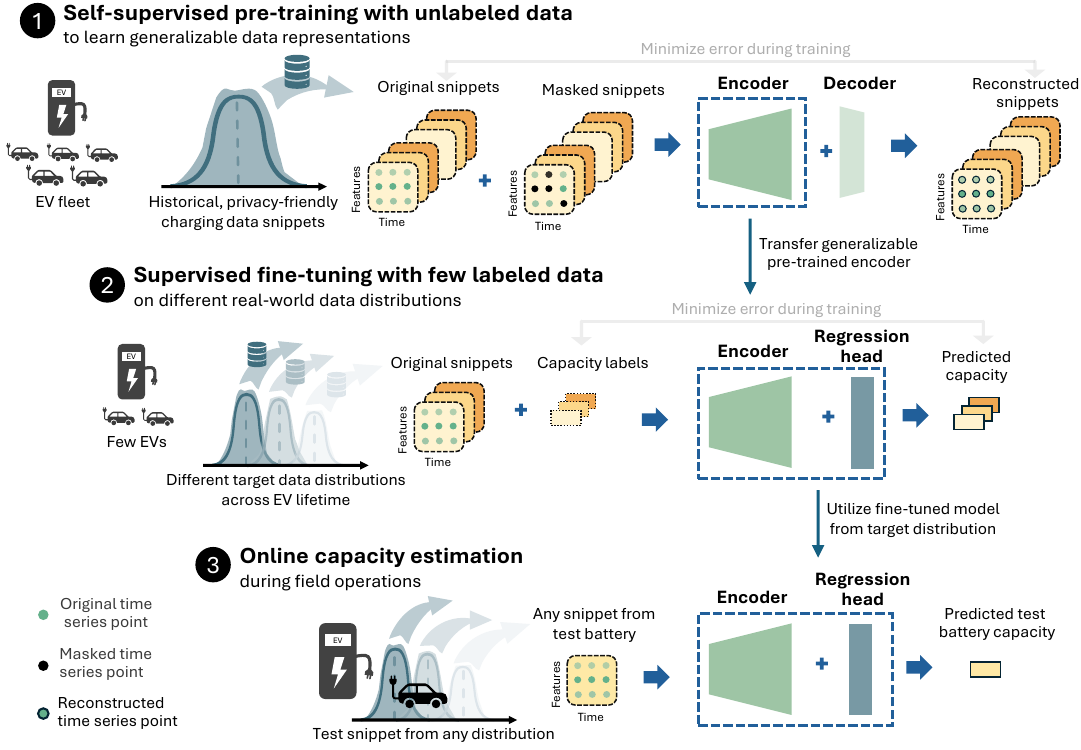}%
        \label{Fig3a}
    }
    \hfill

    \vspace{0.3cm}
    \subfloat[]{%
        \includegraphics[width=0.7\textwidth]{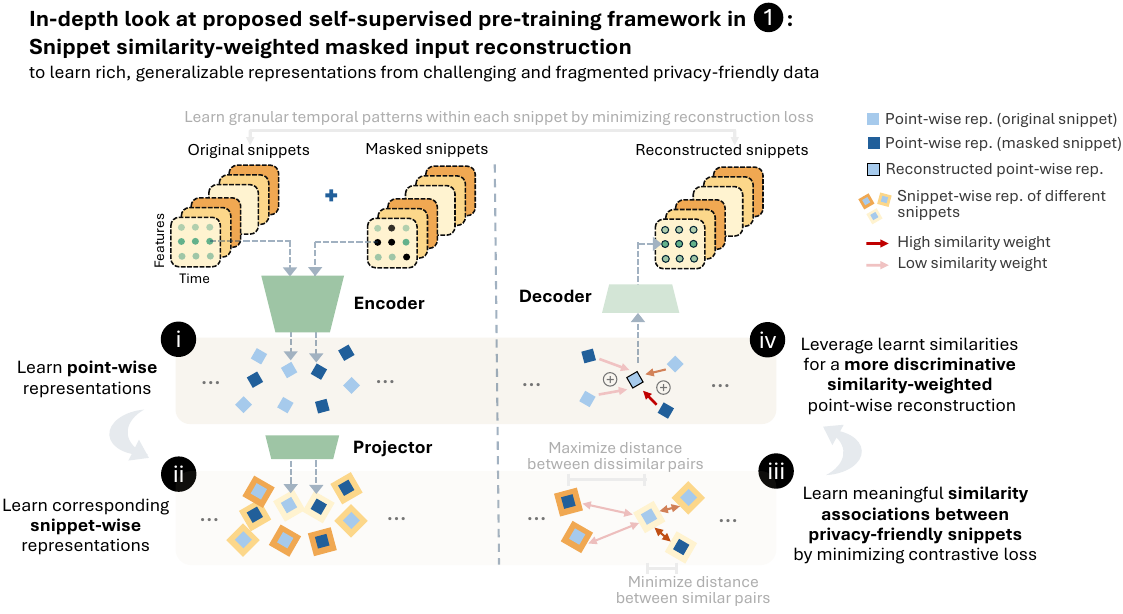}%
        \label{Fig3b}
    }
    \caption{\textbf{(a)} An overview of ``pre-training--fine-tuning'' pipeline for developing a generalizable capacity estimation model, followed by \textbf{(b)} an in-depth look at our proposed pre-training framework. The key innovation is learning meaningful similarities among the fragmented snippets, which enables the model to focus on similarities between the masked and original snippets, while suppressing noise from dissimilar snippets for the masked reconstruction process. With the snippet-wise contrastive learning and subsequent similarity-weighted masked reconstruction, the model captures rich representations of both granular charging patterns within individual snippets and high-level associative relationships across different snippets, significantly enhancing downstream capacity estimation.}
    \label{Fig_overall_framework}
    \vspace{-1em}
\end{figure*}

We detail the capacity estimation framework, with a  focus on the proposed self-supervised pre-training methodology. Fig. \ref{Fig_overall_framework}a provides an overview of the ``pre-training--fine-tuning" pipeline for developing a practically effective capacity estimation model with broad generalization capabilities, even when labeled training data is limited. During pre-training, a well-designed self-supervised learning task (in our case, masked input reconstruction)  trains the encoder to learn generalizable representations from the more readily available unlabeled charging snippets. During fine-tuning, a capacity estimation model for different target distributions can then be quickly trained by fine-tuning the pre-trained encoder on a small set of labeled snippets from the target distribution. The success of this transfer learning, especially with limited labeled data, hinges on the pre-training framework's ability to learn rich, generalizable representations that can be reused across various downstream capacity estimation scenarios. 

Fig. \ref{Fig_overall_framework}b details our proposed pre-training methodology: \textit{snippet similarity-weighted masked input reconstruction}. The basic principle of masked input reconstruction is to reconstruct the masked-out portions of time series as accurately as possible to the original, thereby inducing the learning of meaningful representations in the original time series. However, the higher noise-to-signal ratio in fragmented snippets containing only partial charging data can significantly challenge the masked reconstruction process and hamper effective representation learning. Addressing this critical gap, our pre-training method is specially designed to learn generalizable representations of broad practical applicability, even from less feature-rich and fragmented, privacy-friendly data. The key innovation lies in the snippet-wise similarity learning, which provides meaningful context to fragmented snippets. The learnt similarities are then leveraged for a similarity-weighted, point-wise reconstruction of the masked series, which significantly enhances the masked reconstruction process and, thus, the representation learning needed for the subsequent fine-tuning phase.

\subsection{Self-supervised pre-training: Snippet similarity-weighted masked input reconstruction}\label{ss_pretraining_methodology}

The pre-training method comprises of four sub-components, illustrated step-by-step in Fig. \ref{Fig_overall_framework}b. We elaborate them below.

\subsubsection{Point-wise representation learning}\label{pwrl_methods}

Given $N$ charging snippets $\left\{\mathbf{x}_i\right\}_{i=1}^N$, we first generate a masked snippet $\mathbf{x}_i^* \in \mathbb{R}^{T \times C}$ for each original snippet $\mathbf{x}_i \in \mathbb{R}^{T \times C}$ by randomly masking a portion of the $T$ time points for each battery monitoring variable in $C$. We adopt the well-established geometric masking strategy \cite{zerveas2021transformer} to randomly mask selected temporal subsequences with zeros. 

The original and masked snippets are fed into encoder $f (\cdot)$ to learn high-dimensional representations of each time point, which we henceforth term as \textit{point-wise representations}:
\vspace{-0.2em}
\begin{equation}
f\left(\bigcup_{i=1}^N\left(\left\{\mathbf{x}_i\right\} \cup\left\{\mathbf{x}_i^{*}\right\}\right)\right)= \bigcup_{i=1}^N\left(\left\{\mathbf{p}_i\right\} \cup\left\{\mathbf{p}_i^{*}\right\}\right) = \mathcal{P}
\end{equation}

\noindent where, $\mathbf{p}_i \in \mathbb{R}^{T \times {d}_f}$ and $\mathbf{p}_i^{*} \in \mathbb{R}^{T \times {d}_f}$ are point-wise representations learnt from the original and masked snippets, respectively, and ${d}_f$ is the encoder's hidden dimension. Our proposed encoder is a lightweight yet  efficient, single-layer bidirectional LSTM block, designed to learn important sequential dependencies among time points in both directions.

\subsubsection{Snippet-wise representation learning}\label{swrl_methods}
Next, to explicitly associate the point-wise representations with their respective snippets for the downstream similarity learning, a projector $h (\cdot)$ maps and condenses the point-wise representations into \textit{snippet-wise representations}:
\vspace{-0.1em}
\begin{equation}
h\left(\bigcup_{i=1}^N\left(\left\{\mathbf{p}_i\right\} \cup\left\{\mathbf{p}_i^{*}\right\}\right)\right)= \bigcup_{i=1}^N\left(\left\{\mathbf{s}_i\right\} \cup\left\{\mathbf{s}_i^{*}\right\}\right) = \mathcal{S}
\end{equation}\\
\vspace{-0.1em}
\noindent where, $\mathbf{s}_i \in \mathbb{R}^{1 \times {d}_h}$ and $\mathbf{s}_i^{*} \in \mathbb{R}^{1 \times {d}_h}$ are the snippet-wise representations learnt from the original and masked snippets, respectively, ${d}_h$ is the projector's hidden dimension. The projector is a MLP layer that aggregates the point-wise representations from all $T$ time points in the snippet sequence into a single snippet-wise representation.

\subsubsection{Snippet-wise similarity learning}\label{swsl_methods}
Learning \textit{snippet-wise similarities} is the key feature of our model design for learning effectively from fragmented and short-sequenced privacy-friendly snippets. To successfully reconstruct the original point-wise representation $\mathbf{p}_i$ from its masked counterpart $\mathbf{p}_i^{*}$, the model needs to recognize that both representations originate from the same underlying snippet, and thus their snippet-wise representations $\mathbf{s}_i$ and $\mathbf{s}_i^{*}$ should have high similarity. This is especially crucial when dealing with privacy-friendly, short-sequence snippets, as their relatively higher noise-to-signal ratio can significantly challenge the reconstruction process and, ultimately, hamper the learning of meaningful representations from input snippets.

We leverage contrastive learning \cite{chen2020simple} and propose a new contrastive loss formulation to learn the pair-wise similarities and dissimilarities among the snippet-wise representations in $\mathcal{S}$. The contrastive loss $\mathcal{L}_{c}$ is formalized as: 
\begin{equation}
\mathcal{L}_{c}=-\sum_{\mathbf{s} \in \mathcal{S}}\left(\sum_{\mathbf{s}^{\prime} \in \mathcal{S}^{+}} \log \frac{\exp \left(D_{\mathbf{s}, \mathbf{s}^{\prime}} / \tau\right)}{\sum_{\mathbf{s}^{\prime \prime} \in \mathcal{S} \backslash\{\mathbf{s}\}} \exp \left(D_{\mathbf{s}, \mathbf{s}^{\prime \prime}} / \tau\right)}\right)
\end{equation}

\noindent where, $\mathbf{s}^{\prime} \in \mathcal{S}^{+} \subset \mathcal{S}$ are representations similar to \(\mathbf{s}\), \(D_{\cdot, \cdot}\) is a pairwise cosine similarity value in the similarity matrix $\mathbf{D}$, and \(\tau\) is the temperature parameter controlling the contrast between similar and dissimilar pairs in the softmax normalization above.

Minimizing $\mathcal{L}_{c}$ requires decreasing the cosine distance between similar snippet representations from the original snippets and their masked counterparts, while increasing the distance  between dissimilar snippet representations. This guides the projector $h(.)$ to learn a more discriminative snippet-wise representation space.

\subsubsection{Snippet similarity-weighted point-wise reconstruction}\label{sswpr_methods}

With the learnt snippet-wise similarities, we can now perform a better reconstruction of the $i$-th snippet back in the point-wise representation space $\mathcal{P} $ by taking \textit{similarity-weighted} aggregation of the point-wise representations: 

\begin{equation}
\widehat{\mathbf{p}}_i=\sum_{\mathbf{s}^{\prime} \in \mathcal{S} \backslash\left\{\mathbf{s}_i\right\}} \frac{\exp \left(D_{\mathbf{s}_i, \mathbf{s}^{\prime}} / \tau\right)}{\sum_{\mathbf{s}^{\prime \prime} \in \mathcal{S} \backslash\left\{\mathbf{s}_i\right\}} \exp \left(D_{\mathbf{s}_i, \mathbf{s}^{\prime \prime}} / \tau\right)} \mathbf{p}^{\prime}
\end{equation}

\noindent where, $\mathbf{p}^{\prime}$ is a point-wise representation, $\mathbf{s}^{\prime}$ is its snippet-wise representation, and $\widehat{\mathbf{p}}_i \in \mathbb{R}^{T \times {d}_f}$ is the reconstructed point-wise representation for the original snippet $\mathbf{x}_i \in \mathbb{R}^{T \times C}$. 

Unlike a simple averaging-based reconstruction that assigns equal weights to all representations, including unrelated ones, our similarity-weighted approach enables the model to focus on similarity associations between the masked and original snippet pairs, or even other relevant snippets, while suppressing the noise from dissimilar snippets. Lastly, the reconstructed point-wise representations $\left\{\widehat{\mathbf{p}}_i\right\}_{i=1}^N$ are fed into the decoder $d(\cdot)$ to obtain the reconstructed snippets: 
\begin{equation}
\left\{\widehat{\mathbf{x}}_i\right\}_{i=1}^N=d\left(\left\{\widehat{\mathbf{p}}_i\right\}_{i=1}^N\right)
\end{equation}

\noindent where, $\widehat{\mathbf{x}}_i \in \mathbb{R}^{T \times C}$ is the reconstruction of the original snippet $\mathbf{x}_i \in \mathbb{R}^{T \times C}$. The decoder is just a simple MLP layer. 
The reconstruction loss $\mathcal{L}_{r}$ is formalized as:
\begin{equation}
\mathcal{L}_{r}=\sum_{i=1}^N\left\|\mathbf{x}_i-\widehat{\mathbf{x}}_i\right\|_2^2
\end{equation}

The overall training loss to be minimized for our pre-training task of similarity-weighted masked reconstruction is thus
\vspace{-0.1em}
\begin{equation}
\mathcal{L}_{pretrain} = \mathcal{L}_{r}+\beta \mathcal{L}_{c}
\end{equation}

\noindent where, the weight $\beta$ is automatically tuned based on the homoscedastic uncertainty of each loss term \cite{kendall2018multi}. Incorporating both losses enables our encoder to capture richer data representations---the reconstruction loss aids in learning fine-grained temporal patterns within each snippet, while the contrastive loss promotes learning of high-level associative relationships among the fragmented snippets.

\subsection{Supervised fine-tuning for capacity estimation}\label{sup_ft_methods}
Once pre-training is complete, the pre-trained encoder $f(\cdot)$ can be leveraged for the capacity estimation task by combining it with a new regression head $g(\cdot)$ to estimate capacity $y_i$ of charging snippet $\mathbf{x}_i$. The regression head is just a single fully connected linear layer, ensuring that accurate capacity estimation primarily depends on the effectiveness of the encoder. During supervised fine-tuning, only a small amount of labeled data $\left\{\mathbf{x}_i, y_i\right\}_{i=1}^{\bar{N}}$ with $\bar{N} \ll N$ is needed to train the capacity estimation model $g(f(\cdot))$, supervised by the minimization of the squared error:
\vspace{-0.1em}
\begin{equation}
\mathcal{L}_{finetune}=\sum_{i=1}^{\bar{N}}\left\|y_i-\widehat{y}_i\right\|_2^2
\end{equation}

\section{Experimental results and discussion}
\label{results_discussion}
The model's generalizability is evaluated under three highly challenging out-of-distribution scenarios: (i) adaptability to age-induced distribution shifts across EV lifetime, (ii) transferability across different EV manufacturers, and (iii) robustness to novel test patterns unseen in the limited labeled fine-tuning data. This section discusses the capacity estimation performance and provides insights into the model components driving the performance. 

\begin{table}[t!]
\scriptsize
\caption{Impact of pre-training data on capacity estimation performance, quantified by RMSE (lower is better)}\label{Table_impactofunlabeleddata}
\centering
\begin{tabular}{M{2cm} M{2cm} M{1.3cm} M{1.3cm}}
\toprule
\multirow{2}{2cm}{\centering Setting} &  \multirow{2}{2cm}{\centering Fine-tuning/ Test data distribution} & \multicolumn{2}{c}{Pre-training data} \\ \cmidrule(lr){3-4}
                     &            & EVM1\textbf{D1}  & EVM1$\mathbf{\tilde{D}}$\textbf{1}                \\ 
\midrule
 \centering In-distribution & EVM1\textbf{D1}  & 1.852\tiny±0.339        & \textbf{1.723\tiny±0.308}     \\ 
  \multirow{2}{2cm}{\centering Out-of-distribution} & \rule{0pt}{1.2 em}EVM1\textbf{D2}  & 2.627\tiny±2.504        & \textbf{1.388\tiny±0.252}        \\ 
\rule{0pt}{0.9 em} & EVM1\textbf{D3}   & 1.469\tiny±0.310        & \textbf{1.296\tiny±0.187}        \\ \bottomrule
\end{tabular}
\footnotetext[1]{ To recap, EVM1$\mathbf{\tilde{D}}$\textbf{1} contains all charging snippets, including the unlabeled fast charging snippets, whereas EVM1\textbf{D1} contains only the initially labeled slow-charging snippets, now used without labels for this comparative analysis.}
\footnotetext{Note: Experiments were
repeated with 5 different random seeds. The average RMSE and
standard deviation are reported.}
\vspace{-1em}
\end{table}

\subsection{Overview of unlabeled pre-training data and labeled fine-tuning data}\label{overview_data}
Each charging snippet contains basic battery monitoring variables, such as charging current, voltage, temperature, and is stored as a multivariate time series of length 128. Recalling that only a subset of charging snippets in the dataset have capacity labels, we briefly highlight the differences between the labeled and unlabeled snippets in the last two subplots of Fig. \ref{fig_novel_current_patterns}, focusing on the charging current patterns. Labeled snippets mainly consist of slow-charging patterns, while unlabeled snippets contain a significantly diverse range of charging patterns, including higher fast-charging currents.

\begin{table*}[t!] 
\scriptsize
\centering
\caption{Comparison of capacity estimation performance against state-of-the-art pre-training frameworks for time series}
\label{Table_EVM1bench}
\begin{tabular}{
  M{1.2cm} M{1.3cm}  
  M{0.8cm} M{1.2cm}  
  M{0.8cm} M{1.2cm}  
  M{0.8cm} M{1.2cm}  
  M{0.8cm} M{1.2cm}  
  M{0.8cm} M{1.2cm}  
}
\toprule
\multirow{2}{1.2cm}{\centering Pre-training data} & 
\multirow{2}{1.3cm}{Fine-tuning/ Test data} & 
\multicolumn{2}{c}{Ours} & 
\multicolumn{2}{c}{PITS \cite{lee2024learning} (ICLR'24)} & 
\multicolumn{2}{c}{SimMTM \cite{dong2024simmtm} (NeurIPS'23)} & 
\multicolumn{2}{c}{TST \cite{zerveas2021transformer} (SIGKDD'21)} & 
\multicolumn{2}{c}{TSLANet \cite{eldele2024tslanet} (ICML'24)} \\
\cmidrule(lr){3-4} \cmidrule(lr){5-6} \cmidrule(lr){7-8} \cmidrule(lr){9-10} \cmidrule(lr){11-12}
& & RMSE & MAPE (\%) & RMSE & MAPE (\%) & RMSE & MAPE (\%) & RMSE & MAPE (\%) & RMSE & MAPE (\%)\\
\midrule
\multirow{3}{1.2cm}{\centering EVM1$\mathbf{\tilde{D}}$\textbf{1}} 
& EVM1\textbf{D1} & \textbf{1.396\tiny±0.171} & \textbf{2.640\tiny±0.421} & 1.555\tiny±0.251 & 2.992\tiny±0.547 & 1.560\tiny±0.386 & 3.051\tiny±0.855 & 3.992\tiny±2.326 & 7.698\tiny±4.402 & 2.049\tiny±1.643 & 4.181\tiny±3.772 \\
\rule{0pt}{1.1 em} 
& EVM1\textbf{D2} & \textbf{1.278\tiny±0.175} & \textbf{2.505\tiny±0.389} & 1.452\tiny±0.255 & 2.710\tiny±0.456 & 1.848\tiny±0.659 & 3.994\tiny±1.668 & 2.553\tiny±0.658 & 5.130\tiny±1.351 & 1.989\tiny±0.921 & 4.119\tiny±2.100 \\
\rule{0pt}{1.1 em} 
& EVM1\textbf{D3} & \textbf{1.156\tiny±0.085} & \textbf{2.381\tiny±0.154} & 1.567\tiny±0.592 & 2.818\tiny±0.580 & 1.682\tiny±0.645 & 3.655\tiny±1.572 & 3.297\tiny±0.934 & 6.973\tiny±2.106 & 2.524\tiny±1.965 & 4.918\tiny±4.907 \\
\bottomrule
\end{tabular}
\end{table*}

\begin{table*}[t!]
\scriptsize
\centering
\caption{Comparison of capacity estimation performance when pre-training and fine-tuning on different manufacturers (EVM3 $\rightarrow$ EVM1)}
\label{Table_diff_manu}
\begin{tabular}{
  M{1.2cm} M{1.3cm}  
  M{0.8cm} M{1.2cm}  
  M{0.8cm} M{1.2cm}  
  M{0.8cm} M{1.2cm}  
  M{0.8cm} M{1.2cm}  
  M{0.8cm} M{1.2cm}  
}
\toprule
\multirow{2}{1.2cm}{\centering Pre-training data} & 
\multirow{2}{1.3cm}{Fine-tuning/ Test data} & 
\multicolumn{2}{c}{Ours} & 
\multicolumn{2}{c}{PITS \cite{lee2024learning} (ICLR'24)} & 
\multicolumn{2}{c}{SimMTM \cite{dong2024simmtm} (NeurIPS'23)} & 
\multicolumn{2}{c}{TST \cite{zerveas2021transformer} (SIGKDD'21)} & 
\multicolumn{2}{c}{TSLANet \cite{eldele2024tslanet} (ICML'24)} \\
\cmidrule(lr){3-4} \cmidrule(lr){5-6} \cmidrule(lr){7-8} \cmidrule(lr){9-10} \cmidrule(lr){11-12}
& & RMSE & MAPE (\%) & RMSE & MAPE (\%) & RMSE & MAPE (\%) & RMSE & MAPE (\%) & RMSE & MAPE (\%)\\
\midrule
\multirow{3}{1.6cm}{\centering EVM3$\mathbf{\tilde{D}}$1} 
& EVM1$\mathbf{D1}$ & \textbf{1.402\tiny±0.239} & \textbf{2.671\tiny±0.561} & 1.566\tiny±0.286 & 3.036\tiny±0.620 & 1.613\tiny±0.615 & 3.217\tiny±1.433 & 2.899\tiny±0.975 & 5.455\tiny±1.849 & 3.663\tiny±2.177 & 8.001\tiny±5.471 \\
\rule{0pt}{1.1em}
& EVM1$\mathbf{D2}$ & \textbf{1.344\tiny±0.173} & 2.625\tiny±0.355 & 1.365\tiny±0.165 & \textbf{2.613\tiny±0.325} & 1.903\tiny±0.352 & 4.084\tiny±0.970 & 2.788\tiny±0.500 & 5.477\tiny±1.294 & 4.185\tiny±1.590 & 8.929\tiny±3.600 \\
\rule{0pt}{1.1em}
& EVM1$\mathbf{D3}$ & \textbf{1.206\tiny±0.159} & \textbf{2.429\tiny±0.293} & 1.772\tiny±0.833 & 3.124\tiny±0.824 & 1.651\tiny±0.518 & 3.557\tiny±1.339 & 2.906\tiny±0.649 & 5.954\tiny±1.210 & 3.261\tiny±1.540 & 7.247\tiny±3.945 \\
\bottomrule
\end{tabular}
\end{table*}

For self-supervised pre-training, we only use charging snippets from Distribution 1. Within Distribution 1, we generate two variants for the pre-training data: (i) EVM1$\mathbf{\tilde{D}}$\textbf{1}, which contains all charging snippets, including unlabeled fast-charging snippets, and (ii) EVM1\textbf{D1}, which only contains the initially labeled slow-charging snippets, now used without labels. These variants enable us to assess how harnessing the more diverse unlabeled snippets impacts the downstream capacity estimation performance. For the supervised fine-tuning following pre-training, labeled snippets from any of the Distributions 1 through 3 can be used to train a capacity estimation model optimized for the target distribution. However, unlike prior studies \cite{wang2024lithium, wang2023self, wang2024enhanced}, we utilize only 10\% labeled data to better reflect the real-world scarcity of labeled data. Details are provided  in Table \ref{Table_train_test_split}.

\subsection{Impact of harnessing unlabeled data}\label{impact_unlabeled_data}

The pre-trained encoder's generalizability depends on both the pre-training data quality and the effectiveness of the pre-training framework's design. Before delving into the design, we examine the impact of harnessing the unlabeled data for pre-training. We consider two versions of pre-training data, EVM1$\mathbf{\tilde{D}}$\textbf{1} and EVM1\textbf{D1}, as defined earlier in Section \ref{overview_data}. 
We assess the pre-trained encoder's generalizability through its fine-tuning performance on out-of-distribution capacity estimation scenarios, where clear distribution shifts exist between the target distribution (e.g., Distribution 2 or 3) and the pre-training data from Distribution 1 (recall Fig. \ref{Fig_intro_3_distributions}b).

Table \ref{Table_impactofunlabeleddata} compares the capacity estimation performance of our encoder pre-trained on EVM1\textbf{D1} and EVM1$\mathbf{\tilde{D}}$\textbf{1}, respectively. Pre-training on EVM1$\mathbf{\tilde{D}}$\textbf{1} consistently outperforms EVM1\textbf{D1} across both in-distribution and out-of-distribution test settings, as evidenced by its lower average root mean square errors (RMSE). This outperformance is especially pronounced in out-of-distribution settings, highlighting the importance of leveraging  richer unlabeled charging data for developing a generalizable pre-trained encoder that can withstand data distribution shifts.
Even more remarkably, harnessing the unlabeled data in EVM1$\mathbf{\tilde{D}}$\textbf{1} actually yields \textit{long-term improvements} to capacity estimation performance, as seen from the progressively decreasing RMSEs for the later-stage distributions, EVM1\textbf{D2} and EVM1\textbf{D3}. This benefit stems from the encoder's ability to uncover critical dependencies affecting battery capacity from the unlabeled fast charging operations, which are known to accelerate capacity degradation \cite{zhou2022state}.

\subsection{Adaptability to age-induced distribution shifts}\label{robustness_ageinduced_shift}

Next, we evaluate the effectiveness of our pre-training design by benchmarking our capacity estimation performance against several state-of-the-art time series pre-training frameworks \cite{lee2024learning,dong2024simmtm, zerveas2021transformer, eldele2024tslanet}. These frameworks have demonstrated strong performance across a wide range of domains, including healthcare and IoT. However, they have only been validated on benchmark datasets, and their ability to generalize to more challenging, privacy-friendly field data under realistic data distribution shifts has yet to be investigated. In addition to RMSE, we also report the mean absolute percentage error (MAPE) as a scale-independent measure of relative error.

All models are pre-trained only on EVM1$\mathbf{\tilde{D}}$\textbf{1}, and then fine-tuned on different target distributions (EVM1\textbf{D1} to EVM1\textbf{D3}) across the EV lifetime. Table \ref{Table_EVM1bench} compares the fine-tuned capacity estimation performance of our model against state-of-the-art baselines, PITS \cite{lee2024learning}, SimMTM \cite{dong2024simmtm}, TST \cite{zerveas2021transformer}, and TSLANet \cite{eldele2024tslanet}. Not only does our model achieve the lowest RMSE across all models for both in-distribution settings (EVM1\textbf{D1}) and out-of-distribution settings (EVM1\textbf{D2} and EVM1\textbf{D3}), it is also the only model to maintain a robust capacity estimation performance throughout the distribution shifts from EVM1\textbf{D1} to EVM1\textbf{D3}, as demonstrated by its progressively decreasing RMSE. This is a highly non-trivial achievement for learning from privacy-friendly field data, especially considering the mixed and underwhelming performance delivered by the state-of-the-art baselines. In fact, even on EVM1\textbf{D3}, the distribution with the largest shift from the pretraining data, our model achieves a 26.2\% lower RMSE than the best-performing baseline, PITS.

The suboptimal performance of the baseline pre-training frameworks may be attributed to their masked modeling design, which lacks consideration for more complex time series in field applications. For instance, TST and SimMTM rely on parameter-heavy transformer architectures not inherently suited for time series representation learning \cite{eldele2024tslanet}, while TSLANet and PITS require cumbersome data transformations like patchifying time series and Fourier transforms, which risk losing granular temporal information.

\subsection{Transferability across different EV manufacturers}\label{transferability_ev_manufacturer}

We further evaluate the generalizability of our pre-training framework in an even more demanding transfer learning setting, where the pre-training and fine-tuning data come from two different manufacturers. 
For the pre-training data, we use manufacturer EVM3, which has the smallest number of charging records among the three manufacturers. The number of charging snippets  now available for pre-training is less than half of that from manufacturer EVM1 (see Table \ref{Table_train_test_split}), which allows us to also assess the transferability of our pre-training framework across manufacturers with significantly smaller pre-training data. Due to the smaller dataset, the mileage distribution of the EVM3 charging snippets is mostly from Distribution 1, which we denote as EVM3$\mathbf{\tilde{D}}$\textbf{1} to differentiate the manufacturer. The fine-tuning and test capacity estimation continue to be done on EVM1\textbf{D1} to EVM1\textbf{D3} from manufacturer EVM1, as before.

Table \ref{Table_diff_manu} presents the capacity estimation results for this EVM3-to-EVM1 transfer learning setting. Even when pre-trained on a smaller dataset from a different manufacturer, our model achieves robust capacity estimation on the target distributions from EVM1, surpassing all baselines in RMSE. Although MAPE is marginally higher than PITS on EVM1$\mathbf{D2}$ (2.625\% vs. 2.613\%), this one-off and negligible difference is likely due to data-specific variability rather than any methodological limitation. Crucially, on the most domain-shifted distribution, EVM1\textbf{D3}---affected by both manufacturer-related and age-induced distribution shifts---our model experiences only a 4.3\% performance drop compared to the previous setting of pre-training with same manufacturer. Moreover,  it still achieves a 31.9\% lower RMSE than PITS, the previously best-performing baseline in Table \ref{Table_EVM1bench}.

\begin{figure}[t]
    \centering
    \begin{minipage}[b]{0.31\columnwidth}
    \hspace{1.2cm}
        \includegraphics[height=3cm]{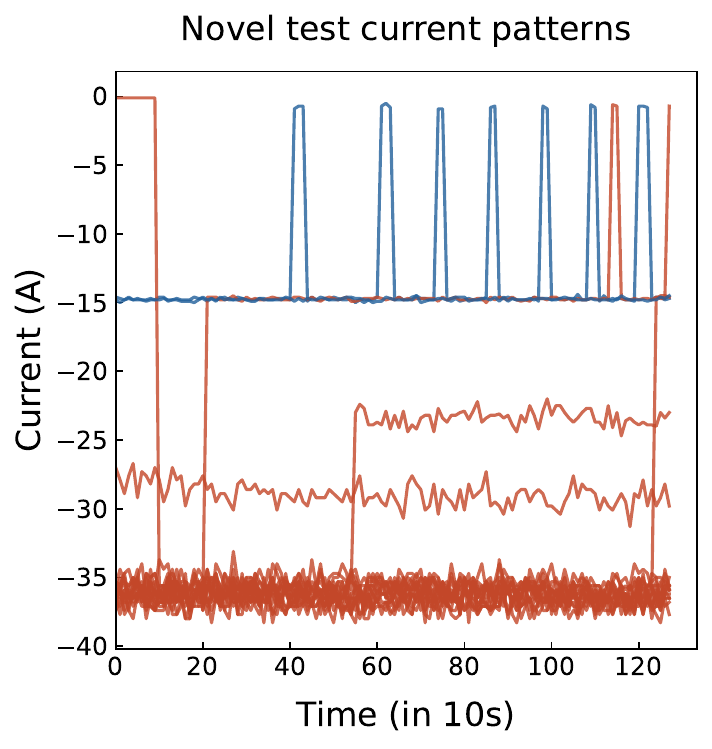}
    \end{minipage}
    \begin{minipage}[b]{0.67\columnwidth}
        \includegraphics[height=3cm]{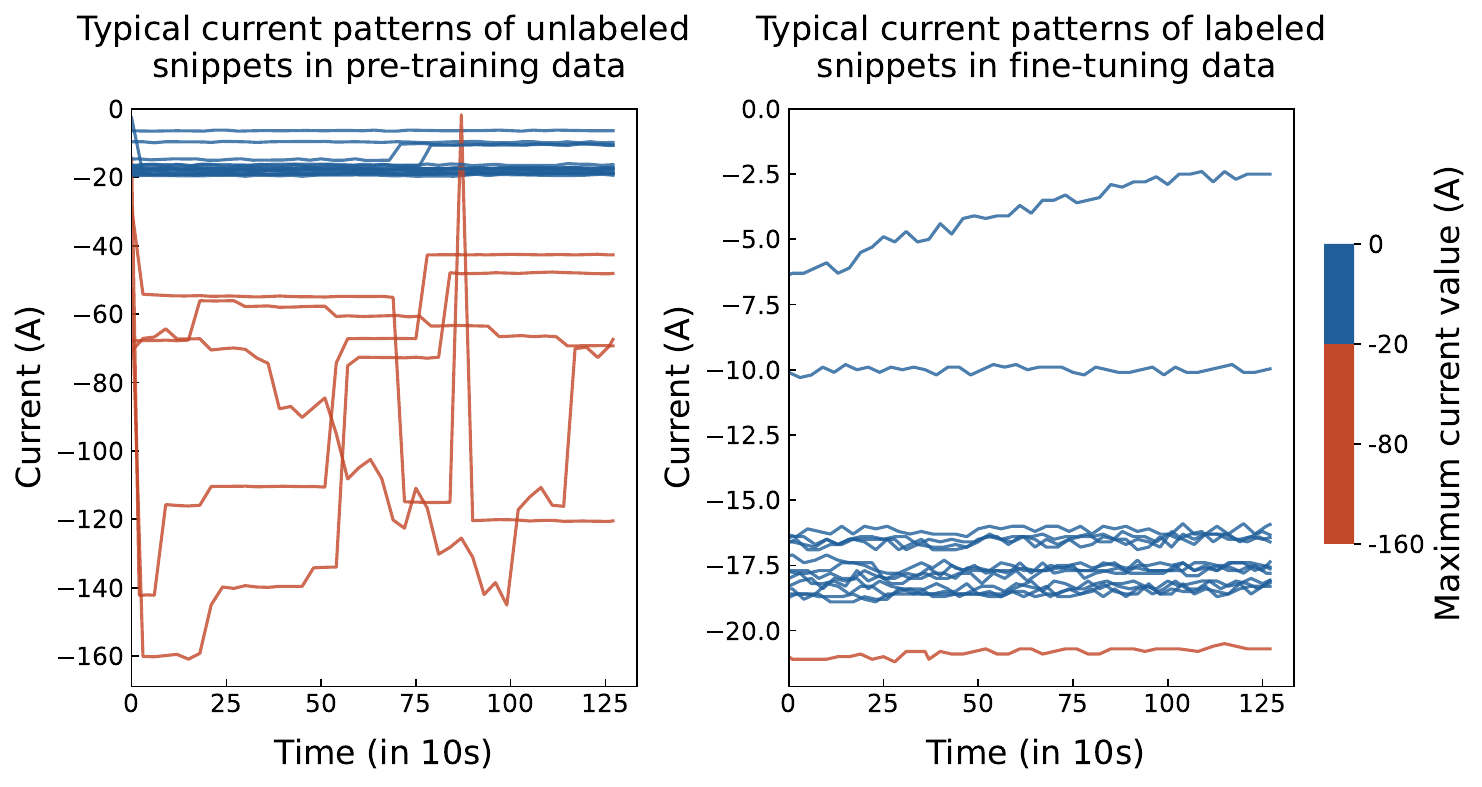}
    \end{minipage}
    \caption{Comparison of some randomly sampled current patterns from the novel test data with typical patterns observed in labeled and unlabeled snippets.}
    \label{fig_novel_current_patterns}
\end{figure}

\begin{figure}[t]
    \centering
    \includegraphics[width=0.55\columnwidth]{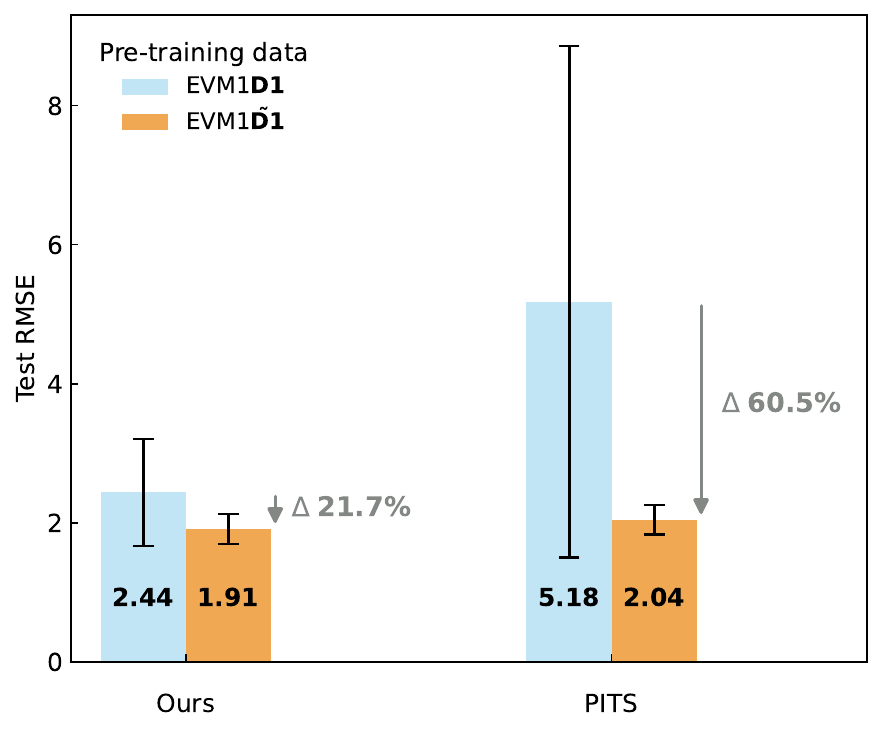}  %
    \vspace{-0.8em}
    \caption{Inference performance on novel test patterns beyond fine-tuning data.}
\vspace{-1em}
\label{fig_rare_pattern}
\end{figure}

\subsection{Robustness to novel test patterns beyond fine-tuning data}\label{robustness_rare_patterns}

We complete our evaluation of the pre-trained encoder’s generalizability by assessing the direct inference performance of our model (without further fine-tuning) on novel test charging patterns unseen in the limited fine-tuning data. To recap, labeled charging snippets available for fine-tuning are mainly from regular, slow-charging
scenarios with current values below 20 A. However, a few rare EVs of the remaining manufacturer, EVM2, have labeled snippets with current values up to 40 A---significantly higher than typical labeled snippets, but still lower than the fast-charging currents in unlabeled snippets (see Fig. \ref{fig_novel_current_patterns}). These novel labeled charging data serve as an insightful test case for evaluating the model’s inference performance on rare yet operationally plausible patterns beyond the limited fine-tuning data.

\begin{table}[t]
\scriptsize
    \centering
    \caption{Ablation study}
    \label{Table_ablation_study}
    \begin{threeparttable}
    \begin{tabular}{M{1.2cm} M{1cm} M{1.9cm} M{1cm} M{1.6cm}}
        \toprule
        Loss components & Pre-train data & Recon. error during pre-training\tnote{1}  & Fine-tune data & Capacity estimation error \\
        \midrule
        \multirow{3}{1.2cm}{ \centering $\mathcal{L}_{r}$ only} 
        & \multirow{3}{1cm}{\centering EVM1$\mathbf{\tilde{D}}$\textbf{1}} & \multirow{3}{2cm}{\centering 0.0439\tiny±0.0025
} & EVM1\textbf{D1} & 1.435\tiny±0.395 \\
        &   & & EVM1\textbf{D2} & 1.317\tiny±0.297 \\
        &   &  & EVM1\textbf{D3} & 1.168\tiny±0.110 \\
        \midrule
        \multirow{3}{1.2cm}{\centering $\mathcal{L}_{c}$ + $\mathcal{L}_{r}$} 
        & \multirow{3}{1cm}{\centering EVM1$\mathbf{\tilde{D}}$\textbf{1}} & \multirow{3}{2cm}{\centering 0.0273\tiny±0.0044}
 & EVM1\textbf{D1} & 1.396\tiny±0.171 \\
        &   &  & EVM1\textbf{D2} & 1.278\tiny±0.175 \\
        &   &  & EVM1\textbf{D3} & 1.156\tiny±0.085 \\
        \bottomrule
    \end{tabular}
\begin{tablenotes}
\item[1]{Reconstruction error is reported as mean squared error, while capacity estimation error is reported as root mean squared error.}
\end{tablenotes} 
\end{threeparttable}       

\end{table}

As the mileage distribution of these test snippets is below 100,000 km and thus falls within Distribution 1, we select the model fine-tuned on EVM1\textbf{D1} for the inference. For pre-training, we revisit the two variants of pre-training data from Distribution 1, EVM1$\mathbf{\tilde{D}}$\textbf{1} and EVM1\textbf{D1}, to also distinguish the impact of the pre-training data versus methodology design on inference performance. Fig. \ref{fig_rare_pattern} compares our model’s direct inference performance on these novel test data (without further fine-tuning) against the strongest baseline from earlier experiments, PITS. Despite the added complexity of the test data, our model remains robust, consistently achieving lower RMSE than PITS, regardless of whether it was pre-trained on the more diverse EVM1$\mathbf{\tilde{D}1}$ or the limited EVM1$\mathbf{D1}$. This robustness is a strong testament to the effectiveness of our pre-training methodology, rather than just a consequence of the scale and quality of the pre-training data. In comparison, not only is PITS's inference performance on novel test patterns worse, but it is also more dependent on the pre-training data quality---as evidenced by the substantial data-driven performance gain ($\Delta 60.5\%$) needed to even achieve a comparable  performance (RMSE = 2.04) to our model (RMSE = 1.91).

\begin{figure}[t]
    \centering
    \includegraphics[width=0.8\columnwidth]{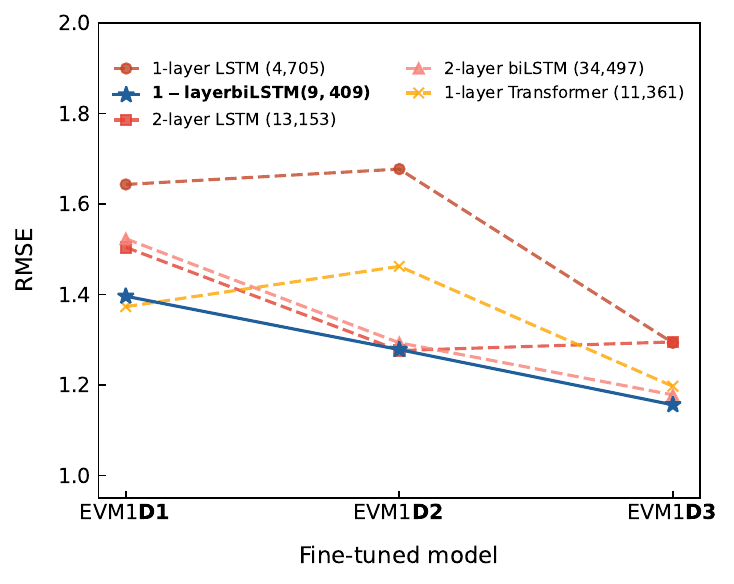} 
    \vspace{-0.2cm}
    \caption{Performance of different encoder architectures (and parameter count).}
 \label{Fig_archi_performance_analysis}
\end{figure}

\begin{figure}[t]
    \centering
    \includegraphics[width=0.98\columnwidth]{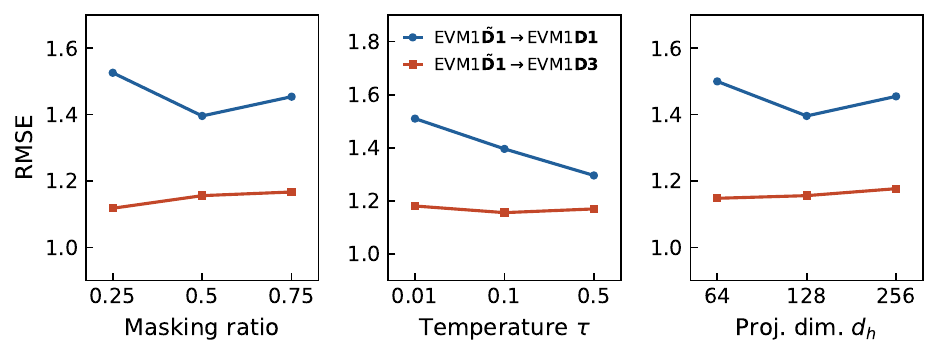} 
    \vspace{-0.5cm}
    \caption{Sensitivity analysis of key hyperparameters on two representative fine-tuning settings: in-distribution EVM1$\mathbf{D1}$ and out-of-distribution EVM1$\mathbf{D3}$.}
    \label{Fig_hyperparam_sens_analysis}
\end{figure}

\vspace{-0.1em}
\subsection{Ablation study and sensitivity analysis}\label{ablation_study}

For the ablation study, we investigate the impact of both the contrastive loss component for similarity learning and the choice of encoder architecture. As shown in Table~\ref{Table_ablation_study}, incorporating similarity learning via contrastive loss not only reduces masked reconstruction error during pre-training, but also improves the accuracy and stability (lower standard deviation) of downstream capacity estimation.

\begin{figure*}[t!]
    \centering

    \subfloat[]{%
        \includegraphics[width=1\textwidth]{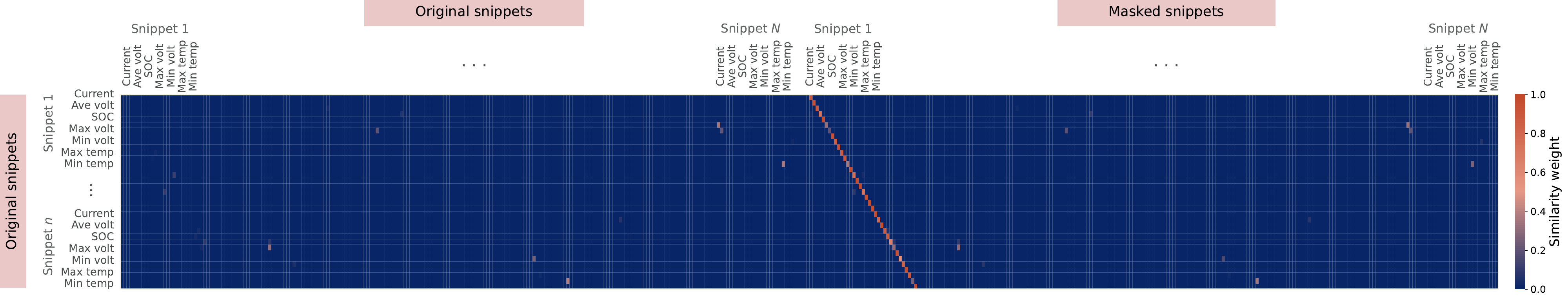}%
        \label{fig:top}
    }

    \vspace{0.1cm} 

    \subfloat[Current]{%
        \begin{minipage}{0.3\textwidth}
            \centering
            \includegraphics[width=0.35\textwidth]{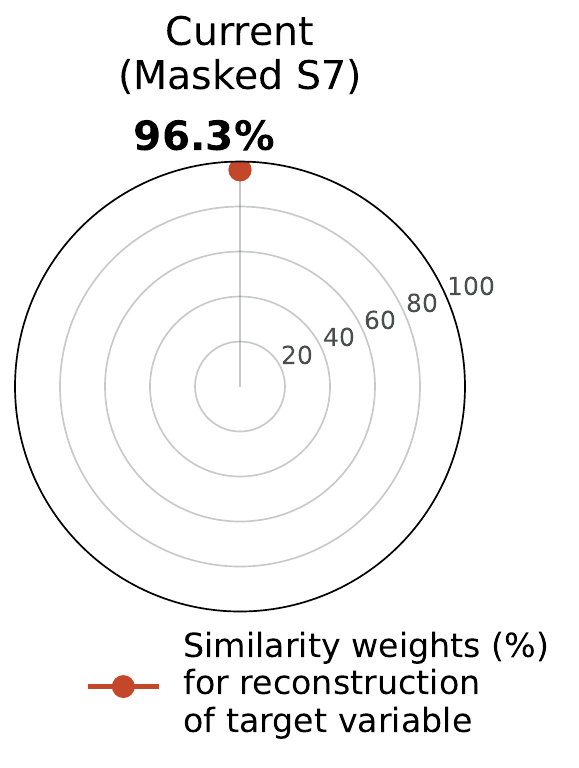}%
            \includegraphics[width=0.48\textwidth]{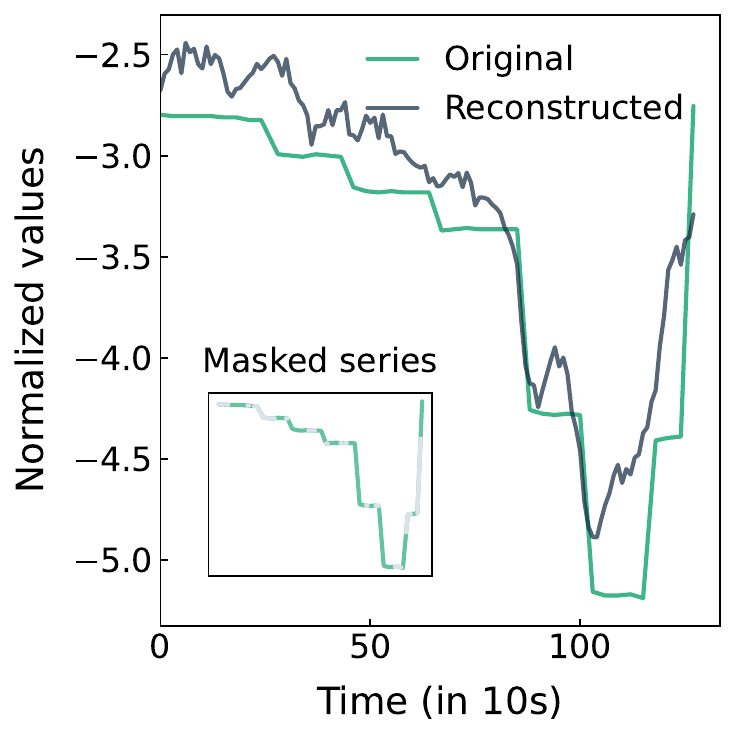}%
        \end{minipage}
    } 
    \hspace{0.02\textwidth}
    \subfloat[Maximum Voltage]{%
        \begin{minipage}{0.3\textwidth}
            \centering
            \includegraphics[width=0.48\textwidth]{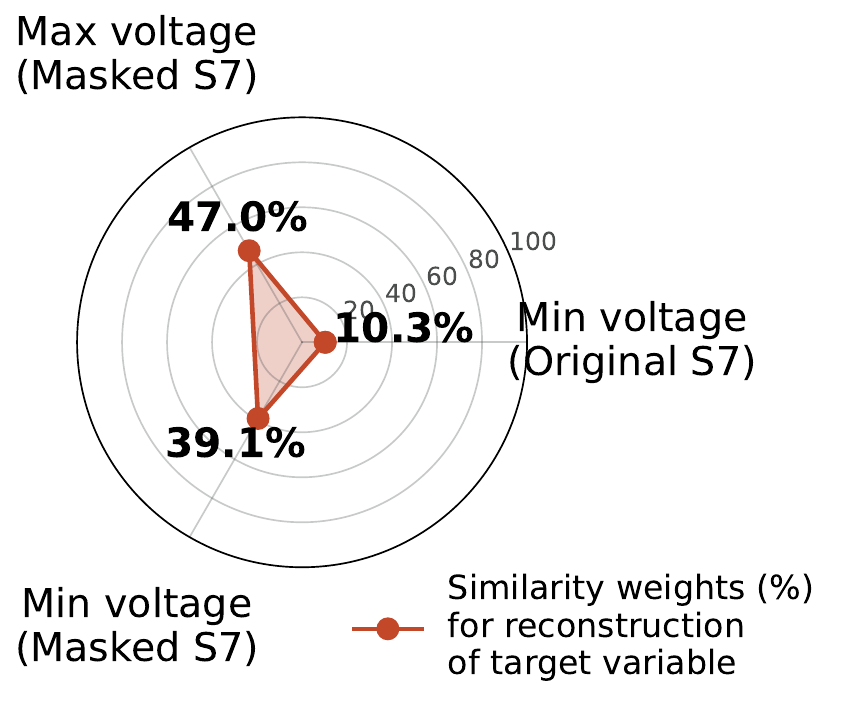}%
            \includegraphics[width=0.48\textwidth]{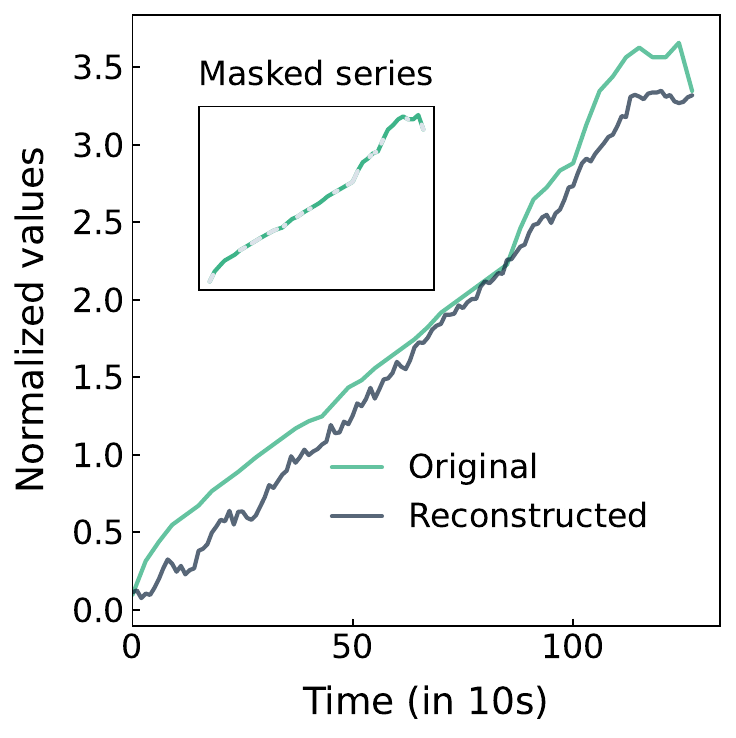}%
        \end{minipage}
    } 
    \hspace{0.02\textwidth}
    \subfloat[Minimum Temperature]{%
        \begin{minipage}{0.3\textwidth}
            \centering
            \includegraphics[width=0.48\textwidth]{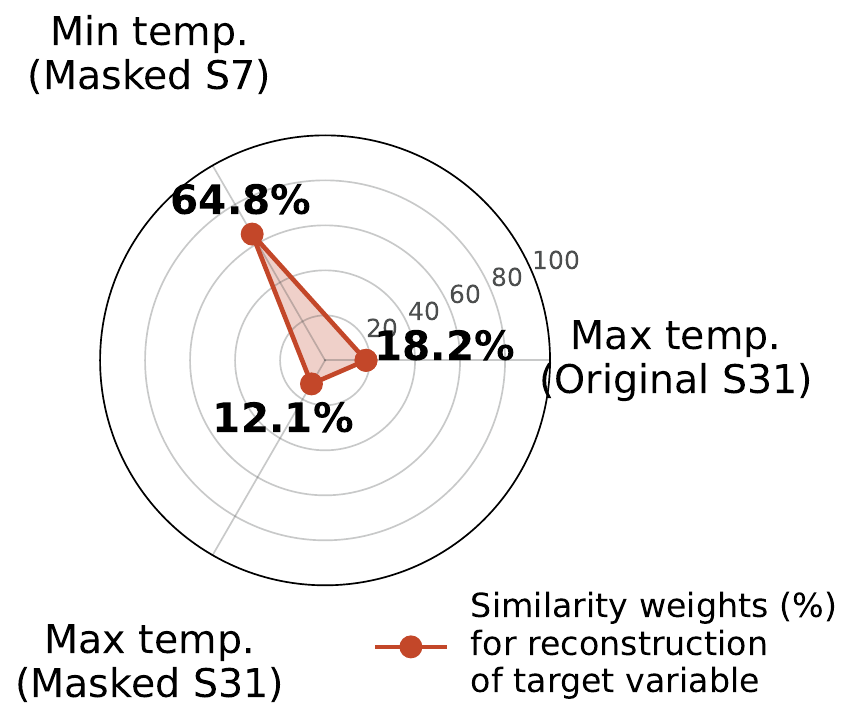}%
            \includegraphics[width=0.48\textwidth]{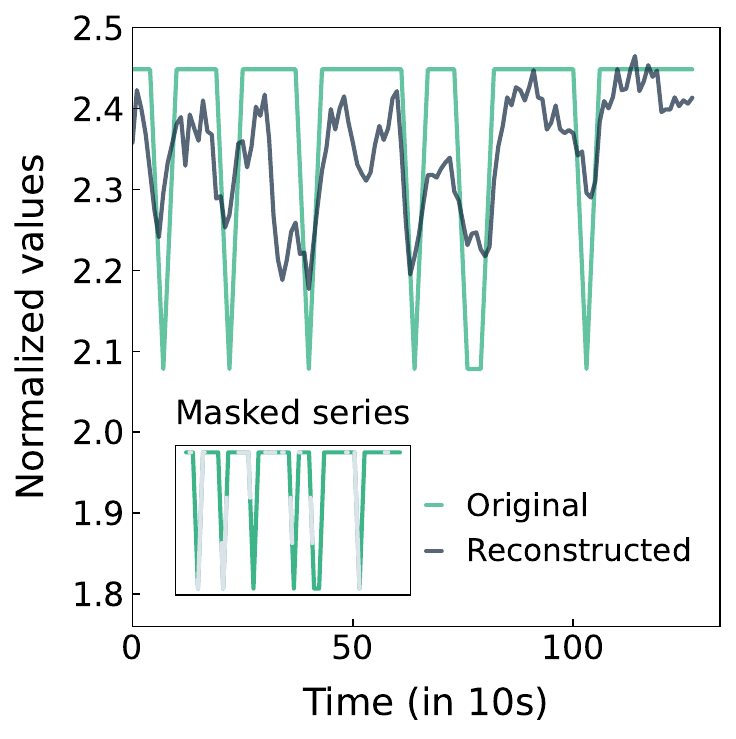}%
        \end{minipage}
    }

    \caption{Understanding the internal mechanism of our model which enables the rich representation learning of both granular temporal patterns within a snippet and high-level similarity associations among different snippets. Taking a high-level view of the similarity matrix, \textbf{(a)} presents a subsample of snippets along the rows and their learned similarity weights to all other snippets along the columns, including both the original and masked versions. \textbf{(b)} to \textbf{(d)} provide a granular view into the similarity-weighted reconstruction process of individual battery charging variables, where S7 denotes Snippet 7, and so on.}
    \label{Fig_knowledge_learnt}
\end{figure*}

Fig. \ref{Fig_archi_performance_analysis} presents the downstream capacity estimation performance and model parameter sizes of several candidate encoder architectures. Compared to transformers or deeper LSTMs, the chosen 1-layer biLSTM encoder achieves the best balance of capacity estimation accuracy, parameter efficiency, and generalization across different fine-tuning scenarios. Its effectiveness stems from the ability to capture temporal dependencies in both past and future directions, which is especially valuable given the limited temporal context in fragmented charging snippets.

We also analyze the model’s sensitivity to key hyperparameters such as  masking ratio, projector dimension size ${d}_h$, and contrastive learning temperature constant $\tau$. Fig. \ref{Fig_hyperparam_sens_analysis} reports the capacity estimation performance on two representative fine-tuning settings: in-distribution EVM1$\mathbf{D1}$ and out-of-distribution EVM1$\mathbf{D3}$. A masking ratio of 0.5 strikes a good balance between reconstruction difficulty and effective representation learning, yielding strong capacity estimation performance in both in-distribution and out-of-distribution settings. Increasing the ratio beyond 0.5 to 0.75 tends to degrade performance, likely due to the greater complexity of reconstructing heavily masked field data. Next, the projector dimension ${d}_h$ governs the expressiveness of the snippet-wise representation space used for computing contrastive similarity. Strong capacity estimation performance is achieved in both in-distribution and out-of-distribution settings with ${d}_h=128$. Increasing ${d}_h$ beyond 128 to 256 leads to performance deterioration, likely due to overparameterization of the projector relative to the  input features. Finally, the temperature parameter $\tau$ in the contrastive loss controls the sharpness of the similarity distribution over snippet-wise representation pairs. If $\tau$ is too low (e.g., 0.01 as shown in Fig. \ref{Fig_hyperparam_sens_analysis}), similarity learning may be constrained to only the most similar pairs, such as masked versions of the same snippet. If too high, the learnt similarities may not be meaningfully discriminative. Reasonably ranged values like 0.1 and 0.5 yield stronger capacity estimation, with $\tau = 0.1$ providing a favorable balance across both in-distribution and out-of-distribution settings.

\subsection{Interpretability analysis}\label{inter_analysis}
We investigate the internal learning mechanism that underpins the rich representation learning and generalizability of our framework. The key feature of our \textit{snippet similarity-weighted masked reconstruction} is its ability to capture both fine-grained temporal patterns within a snippet and high-level similarity associations across snippets. We begin with a high-level examination of the learnt similarity weights used during masked reconstruction. Due to the large size of the similarity matrix, Fig. \ref{Fig_knowledge_learnt}a only presents a subset of snippets and their learnt similarity weights to other snippets, including both the original and masked versions.  Each row represents the similarity weights contributing to the reconstruction of a single battery monitoring time series variable (e.g., current) in a given snippet. As expected, the highest weights typically correspond to the masked version of the same snippet (seen as offset diagonals). However, the presence of large similarity weights at some non-offset positions indicates that the model also leverages similarities from other relevant variables and snippets to improve reconstruction.

Figs. \ref{Fig_knowledge_learnt}b–d offer a granular view of the reconstruction process for individual battery time series variables from selected snippets. The radial plots (left) highlight the battery variables and snippets with the highest similarity weights contributing to the target variable’s reconstruction. The reconstruction plots (right) depict the corresponding reconstruction performance for the target variable. As evident from these plots, reconstructing field data is particularly challenging due to the noisy and erratic nature of signals in field applications. Thus, a successful reconstruction relies on effective similarity learning from appropriate snippets---whether from the masked version of the target snippet (Fig. \ref{Fig_knowledge_learnt}b), from other correlated variables within the same snippet (Fig. \ref{Fig_knowledge_learnt}c), or even from entirely different but contextually relevant snippets (Fig. \ref{Fig_knowledge_learnt}d).

\subsection{Computational complexity analysis}
\label{comp_complexity}

The main source of computational complexity in our pre-training method is the pairwise similarity computation. The batched similarity matrix $\mathbf{D}$ has size $2M \times 2M$, where $M = b \cdot C$, with $b$ as the batch size and $C$, to recap, the number of battery monitoring variables.

Therefore, the memory complexity of storing the similarity matrix is $\mathcal{O}(b^2 C^2)$, and the time complexity of the similarity computation is $\mathcal{O}(b^2 C^3)$. As a large batch size can increase training time, we adopt $b = 32$ to balance computational efficiency and representational learning adequacy. Moreover, the number of input feature variables in the privacy-preserving charging data is inherently low, with $C = 7$. Consequently, the similarity matrix computation takes approximately just $3.03 \times 10^{-4}$ seconds per batch-per epoch, with a peak memory usage of 0.99 MB on a NVIDIA RTX 3070 GPU.

In this work, the scale of the similarity matrix computation is manageable, as our pre-training method is designed to effectively leverage  less feature-rich, privacy-preserving data. However, optimizations such as top-$K$ neighbor truncation \cite{dwibedi2021little} or a modified memory bank \cite{bulat2021improving} could also be explored in future iterations.

\subsection{Implementation details}
\label{implementation_details}

We summarize the key implementation details here. A 70\%–10\%–20\% split of vehicles is adopted for pre-training, validation, and testing. For fine-tuning, only 10\% of vehicles within the pre-training split are used. Snippet-level normalization is performed using a range-adjusted z-score normalization, as recommended for this dataset in \cite{he2022evbattery}.

For time series masking, we apply geometric masking \cite{zerveas2021transformer} with a masking ratio of 0.5, randomly replacing temporal subsequences with zeros. For contrastive learning, $\tau = 0.1$. Batch size is set to 32. The model architecture consists of a single-layer biLSTM encoder with  ${d}_f = 32$, an MLP projector with ${d}_h = 128$, and a linear regression head with output dimension 1. 

Model training is performed with the Adam optimizer. The learning rate is scheduled with a maximum initial value of 0.01. Pre-training is conducted for 50 epochs, averaging around 83.06 seconds per epoch. Fine-tuning is conducted for 200 epochs with early stopping, averaging around 0.86 seconds per epoch. The inference time is negligible, averaging approximately $3.80 \times 10^{-5}$ seconds per test snippet.
 
Experiments were conducted on a NVIDIA RTX 3070 (8GB) GPU. Each experiment was repeated with 5 different random seeds to ensure robustness of model performance. The average model performance and standard deviation are reported. For benchmark models, we follow hyperparameters recommended in the respective papers and official code, applying additional tuning as needed for our dataset.

\section{Conclusion}
\label{conclusion}

To realize the potential of DL for practical EV adoption, generalizable models capable of adapting to real-world distribution shifts are needed. However, no prior studies have delved deeply into model generalizability and rigorously tested their capacity estimation models on realistic out-of-distribution scenarios. Compounded with the challenges of privacy and data labeling costs, there is a need to rethink the reliance on extensive, well-labeled training data to achieve model generalizability. In this work, we developed a first-of-its-kind capacity estimation model based on self-supervised learning, designed to learn rich, generalizable representations even from less feature-rich and fragmented privacy-friendly data. With our snippet-wise contrastive learning and subsequent similarity-weighted masked reconstruction, we can learn rich representations of both granular charging patterns within individual snippets \textit{and} high-level associative relationships across different snippets. Bolstered by the rich representation learning, our capacity estimation framework achieved new state-of-the-art performance across multiple out-of-distribution settings, with only 10\% labeled fine-tuning data.

We encourage the research community to build on this new research direction and prioritize the development of generalizable, future-ready prognostics models that are labeled data-efficient and privacy-friendly. Another future area of interest is integrating difficulty-aware uncertainty quantification into the framework to provide prediction confidence as an added reliability measure, particularly under extreme distribution shifts. Beyond battery capacity estimation, as our pre-training framework is generalizable and domain-agnostic, it can also be adapted to other applications where learning from fragmented, privacy-protected sensor data is crucial (e.g., autonomous vehicles, smart manufacturing).

\appendices
\section{}
\label{app_a}
%
\renewcommand{\thetable}{A.\arabic{table}} 
\setcounter{table}{0} 

\vspace{-1em}
\begin{table}[h]
\scriptsize
\caption{Overview of data splits used for pre-training and fine-tuning}\label{Table_train_test_split}
\centering
\begin{threeparttable}
\begin{tabular}{M{1.2cm} M{1.5cm} M{1.2cm} M{1.2cm} M{1.2cm}}
\toprule
\multirow{2}{*}{Task} & \multirow{2}{1.5cm}{\centering Data distribution} & Train & Validation & Test  \\
\cmidrule(r){3-5}
     &           & \multicolumn{3}{c}{Number of vehicles\tnote{1} / Number of snippets\tnote{2}} \\
\midrule

 \multirow{3}{*}{Pre-training} & EVM1\textbf{D1} & 23 /  54,234  & 3 / 6,007 & NA \\
 & EVM1$\mathbf{\tilde{D}}$\textbf{1}     & 23 /  69,354 & 3 /  7,598 & NA \\
  & EVM3$\mathbf{\tilde{D}}$\textbf{1}     & 5 /  28,937 & 1 / 7,350 & NA \\

\midrule

 \multirow{3}{*}{Fine-tuning} & EVM1\textbf{D1} & 2 / 4,663 & 3 / 6,007 & 6 /  13,584 \\
 
  & EVM1\textbf{D2} &  2 /  3,472 & 3 / 6,388 & 7 /  14,170  \\

                & EVM1\textbf{D3} & 2 /  4,582 & 3 /  4,553 & 7 /  11,333 \\   
                        
\bottomrule

\end{tabular}

\begin{tablenotes}
\item[1]{We adopt a 70\%–10\%–20\% split of vehicles for pre-training, validation, and testing. For fine-tuning, only 10\% of vehicles within the pre-training split (i.e., two vehicles’ worth of data) are used. Each vehicle is uniquely assigned to exactly one split, ensuring that the train–validation–test splits are strictly disjoint with no cross-split information leakage.}

\item[2]{For brevity, we report only the average number of vehicles and charging snippets across five randomly generated train-validation-test splits used in the experiments.}

\end{tablenotes} 
\end{threeparttable}       
\end{table}

\bibliography{mybib_x}
\bibliographystyle{IEEEtran}

\end{document}